# REx: An Efficient Rule Generator

S. M. Kamruzzaman

Department of Computer Science and Engineering
Manarat International University, Dhaka, Bangladesh
E-MAIL: smzaman@gmail.com, smk.cse@manarat.ac.bd

## ABSTRACT

This paper describes an efficient algorithm REx for generating symbolic rules from artificial neural network (ANN). Classification rules are sought in many areas from automatic knowledge acquisition to data mining and ANN rule extraction. This is because classification rules possess some attractive features. They are explicit, understandable and verifiable by domain experts, and can be modified, extended and passed on as modular knowledge. REx exploits the first order information in the data and finds shortest sufficient conditions for a rule of a class that can differentiate it from patterns of other classes. It can generate concise and perfect rules in the sense that the error rate of the rules is not worse than the inconsistency rate found in the original data. An important feature of rule extraction algorithm, REx, is its recursive nature. They are concise, comprehensible, order insensitive and do not involve any weight values. Extensive experimental studies on several benchmark classification problems, such as breast cancer, iris, season, and golf-playing, demonstrate the effectiveness of the proposed approach with good generalization ability.

## 1. INTRODUCTION

ANNs have been successfully applied in a variety of problem domains [1]. In many applications, it is highly desirable to extract symbolic classification rules from these networks. Unlike a collection of weights, symbolic rules can be easily interpreted and verified by human experts. They can also provide new insides into the application problems and the corresponding data.

While the predictive accuracy obtained by ANNs is often higher than that of other methods or human experts, it is generally difficult to understand how ANNs arrive at a particular conclusion due to the complexity of the ANNs architectures [2]. It is often said that an ANN is practically a "black box". Even for an ANN with only single hidden layer, it is generally impossible to explain why a particular pattern is classified as a member of one class and another pattern as a member of another class [3].

This paper proposes an efficient algorithm REx for generating symbolic rules from ANN. A three-phase training algorithm REANN is proposed for backpropagation learning. In the first phase, appropriate network architecture is determined using constructive and pruning algorithm. In the second phase, the continuous activation values of the hidden nodes are discretized by using an efficient heuristic clustering algorithm. And finally in the third phase, rules are extracted by examining the discretized activation values of the hidden nodes using the rule extraction algorithm REx.

## 2. THE REANN ALGORITHM

The aim of this section is to introduce the REANN algorithm for understanding how an ANN solves a given problem. The major steps of REANN are summarized in Fig. 1 and explained further as follows:

**Step 1** Create an initial ANN architecture. The initial architecture has three layers, i.e. an input, an output, and a hidden layer. The number of nodes in the input and output layers is the same as the number of inputs and outputs of the problem. Initially, the hidden layer contains only one node. Randomly initialize all connection weights within a certain small range. The number of nodes in the hidden layer is automatically determined by using a basic constructive algorithm. Remove redundant input nodes and connections by using a basic pruning algorithm.

**Step 2** Discretize the continuous outputs of hidden nodes by using an efficient heuristic clustering algorithm.
**Step 3** Generate rules that map the inputs and outputs relationships using rule extraction algorithm REx.

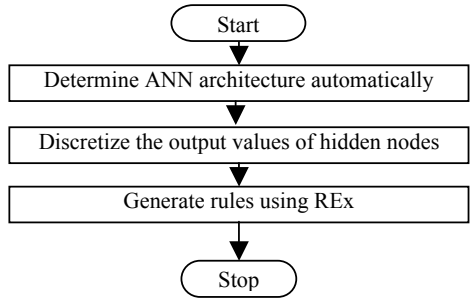

**Fig. 1:** Flow chart of the REANN algorithm.

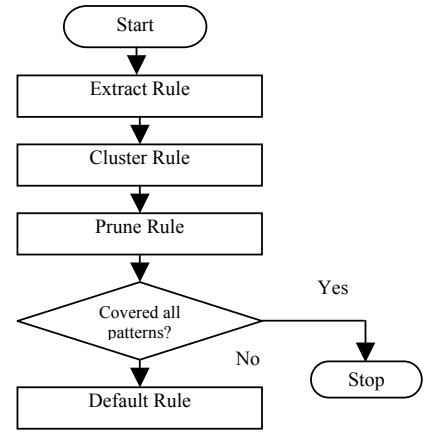

**Fig. 2:** Flow chart of the REx algorithm.

## 3. RULE EXTRACTION ALGORITHM (REx)

Classification rules are sought in many areas from automatic knowledge acquisition [5] [6] to data mining [7] and ANN rule extraction. This is because classification rules possess some attractive features. They are explicit, understandable and verifiable by domain experts, and can be modified, extended and passed on as modular knowledge. The REx is composed of three major functions:
  i) Rule Extraction: this function iteratively generates shortest rules and remove/marks the patterns covered by each rule until all patterns are covered by the rules.
  ii) Rule Clustering: rules are clustered in terms of their class levels and
  iii) Rule Pruning: redundant or more specific rules in each cluster are removed.

A default rule should be chosen to accommodate possible unclassifiable patterns. If rules are clustered, the choice of the default rule is based on clusters of rules.

The steps of the Rule Extraction (REx) algorithm are summarized in Fig. 2, which are explained further as follows:
**Step 1** Extract Rule:
i=0; while (data is NOT empty/marked){
generate Ri to cover the current pattern and differentiate it from patterns in other categories;
remove/mark all patterns covered by Ri ; i++}
**Step 2** Cluster Rule:
Cluster rules according to their class levels. Rules generated in Step 1 are grouped in terms of their class levels. In each rule cluster, redundant rules are eliminated; specific rules are replaced by more general rules.

**Step 3** Prune Rule:
  replace specific rules with more general ones;
  remove noise rules;
  eliminate redundant rules;
**Step 4** Check whether all patterns are covered by any rules. If yes then stop, otherwise continue.
**Step 5** Determine a default rule:
A default rule is chosen when no rule can be applied to a pattern.

REx exploits the first order information in the data and finds shortest sufficient conditions for a rule of a class that can differentiate it from patterns of other classes. It can generate concise and perfect rules in the sense that the error rate of the rules is not worse than the inconsistency rate found in the original data. The novelty of REx is that the rule generated by it is order insensitive, i.e, the rules need not be required to fire sequentially.

## 4. EXPERIMENTAL STUDIES

This section evaluates the performance of REx on several well-known benchmark classification problems. These are the breast cancer, iris, season, and golf playing. They are widely used in machine learning and ANN research. The data sets representing all the problems were real world data and obtained from the UCI machine learning benchmark repository [8]. The characteristics of the data sets are summarized in Table 1.

**Table 1:** Characteristics of data sets.

| Data Sets | No. of Examples | Input Attributes | Output Classes |
|---|---|---|---|
| Breast Cancer | 699 | 9 | 2 |
| Iris | 150 | 4 | 3 |
| Season | 11 | 3 | 4 |
| Golf Playing | 14 | 4 | 2 |

## 4.1 Extracted Rules

Table 2 shows the number of rules extracted by REx and the accuracy of the rules. In most of the cases REx produces fewer rules with better accuracy. It was observed that two to three rules were sufficient to solve the problems. The accuracy was 100% for season and golf playing problems, because of the lower number of examples.

**Table 2:** Number of rules and rules accuracy.

| Data Sets | No. of Extracted Rules | Rules Accuracy |
|---|---|---|
| Breast Cancer | 2 | 96.28 % |
| Iris | 3 | 97.33 % |
| Season | 4 | 100 % |
| Golf Playing | 3 | 100 % |

The number of rules extracted by REx and the accuracy of the rules were described in Table 2. But the visualization of the rules in terms of the original attributes was not discussed. The following subsections discussed the rules extracted by REx in terms of the original attributes. The number of conditions per rule and the number of rules extracted were also visualized here.

**Breast Cancer Problem**

Rule 1: If Clump thickness ($A_1$) <= 0.6 and Bare nuclei ($A_6$) <= 0.5 and Mitosis ($A_9$) <= 0.3, then benign
Default Rule: malignant.

**Iris Problem**

Rule 1: If Petal-length (A3) <= 1.9 then Iris setosa
Rule 2: If Petal-length (A3) <= 4.9 and Petal-width (A4) <= 1.6 then Iris versicolor
Default Rule: Iris virginica.

**Season Problem**

Rule 1: If Tree (A2) = yellow then autumn
Rule 2: If Tree (A2) = leafless then autumn
Rule 3: If Temperature (A3) = low then winter
Rule 4: If Temperature (A3) = high then summer
Default Rule: spring.

**Golf Playing Problem**

Rule 1: If Outlook (A1) = sunny and Humidity >=85 then don't play
Rule 2: Outlook (A1) = rainy and Wind= strong then don't play
Default Rule: play.

## 5. COMPARISON

This section compares experimental results of REx with the results of other works. The primary aim of this work is not to exhaustively compare REx with all other works, but to evaluate REx in order to gain a deeper understanding of rule extraction.

**Table 3:** Performance comparison of REx with other algorithms for **breast cancer** problem.

| Data Set | Feature | REx | NN RULES | DT RULES | C4.5 | NN-C4.5 | OC1 | CART |
|---|---|---|---|---|---|---|---|---|
| Breast Cancer | No. of Rules | 2 | 4 | 7 | - | - | - | - |
| | Avg. No. of Conditions | 3 | 3 | 1.75 | - | - | - | - |
| | Accuracy % | 96.28 | 96 | 95.5 | 95.3 | 96.1 | 94.99 | 94.71 |

**Table 4:** Performance comparison of REx with other algorithms for **iris** problem.

| Data Set | Feature | REx | NN RULES | DT RULES | BIO RE | Partial RE | Full RE |
|---|---|---|---|---|---|---|---|
| Iris | No. of Rules | 3 | 3 | 4 | 4 | 6 | 3 |
| | Avg. No. of Conditions | 1 | 1 | 1 | 3 | 3 | 2 |
| | Accuracy % | 98.67 | 97.33 | 94.67 | 78.67 | 78.67 | 97.33 |

**Table 5:** Performance comparison of REx with other algorithms for **season** problem.

| Data set | Feature | REx | RULES | X2R |
|---|---|---|---|---|
| Season | No. of Rules | 5 | 7 | 6 |
| | Avg. No. of Conditions | 1 | 2 | 1 |
| | Accuracy % | 100.0 | 100.0 | 100.0 |

**Table 6:** Performance comparison of REx with other algorithms for **golf playing** problem.

| Data set | Feature | REx | RULES | RULES-2 | X2R |
|---|---|---|---|---|---|
| Golf Playing | No. of Rules | 3 | 8 | 14 | 3 |
| | Avg. No. of Conditions | 2 | 2 | 2 | 2 |
| | Accuracy % | 100.0 | 100.0 | 100.0 | 100.0 |

Table 3 compares REx results of breast cancer problem with those produced by NN RULES [2], DT RULES [2], C4.5 [6], NN-C4.5 [9], OC1 [9], and CART [10] algorithms. REx achieved best performance although NN RULES was closest second. But number of rules extracted by REx are 2 whereas these were 4 for NN RULES.

Table 4 compares REx results of iris problem with those produced by NN RULES, DT RULES, BIO RE [11], Partial RE [11], and Full RE [11] algorithms. REx achieved 98.67% accuracy although NN RULES was closest second with 97.33% accuracy. Here number of rules extracted by REx and NN RULES are equal.

Table 5 compares REx results of season problem with those produced by RULES [12] and X2R [4]. All three algorithms achieved 100% accuracy. This is possible because the number of examples is low. Number of extracted rules by REx are 5 whereas these were 7 for RULES and 6 for X2R.

Table 6 compares REx results of golf playing problem with those produced by RULES, RULES-2 [13], and X2R. All four algorithms achieved 100% accuracy because the lower number of examples. Number of extracted rules by REx are 3 whereas these were 8 for RULES and 14 for RULES-2.

### 6. CONCLUSIONS

This work is an attempted to open up these black boxes by extracting symbolic rules from it through the proposed efficient rule extraction algorithm REx. The REx algorithm can extract concise rules from standard feedforward ANN. An important feature of rule extraction algorithm, REx, is its recursive nature. They are concise, comprehensible, order insensitive and do not involve any weight values. The accuracy of the rules from a pruned network is as high as the accuracy of the fully connected network.

Extensive experiments have been carried out in this study to evaluate how well REx performed on four benchmark classification problems in ANNs including breast cancer, iris, season, and golf playing in comparison with other algorithms. In almost all cases, REx outperformed the others. With the rules extracted by the method introduced here, ANNs should no longer be regarded as black boxes.

### REFERENCES

[1] R. Setiono and W. K. Leow, " FERNN: An algorithm for fast extraction of rules from neural networks," Applied Intelligence, vol. 12, 2000, pp. 15-25.

[2] R. Setiono and H. Liu, "Symbolic presentation of neural networks," IEEE Computer, March 1996, pp. 71-77.

[3] R. Setiono, "Extracting rules from neural networks by pruning and hidden-unit node splitting," Neural Computation, vol. 9, 1997, pp. 205-225.

[4] H. Liu and S. T. Tan, "X2R: A fast rule generator," Proceedings of IEEE International Conference on Systems, Man and Cybernetics, Vancouver, CA, 1995.

[5] Han Jiawei, Micheline Kamber, "Data Mining: Concepts and Techniques," Morgan Kaufmann Publisher: CA, 2001.

[6] J. R. Quinlan, "C4.5: Programs for Machine Learning," Morgan Kaufmann, San Mateo, CA, 1993.

[7] R. Agrawal, T. Imielinski, and A. Swami, "Database mining: A performance perspective," IEEE Transactions on Knowledge and Data Engineering, vol. 5, pp. 914-925, 1993.

[8] C. Blake, E. Keogh, and C. J. Merz, "UCI repository of of machine learning databases [http://www.ics.uci.edu/~mlearn/MLRepository.htm]," Department of Information and Computer Science, University of California, Irvine, CA, 1998.

[9] R. Setiono, "Techniques for extracting rules from artificial neural networks," Plenary lecture presented at the 5th International Conference on Soft Computing and Information Systems, Iizuka, Japan, October 1998.

[10] L. Breiman, J. Friedman, R. Olshen, and C. Stone, "Classification and Regression Trees," Wadsworth and Brooks, Monterey, CA, 1984.

[11] I. Taha and J. Ghosh, "Three techniques for extracting rules from feedforward networks," Intelligent Engineering Systems Through Artificial Neural Networks, vol. 6, pp. 23-28, ASME Press, St. Louis, 1996.

[12] D. T. Pham and M. S. Aksoy, "Rules: A simple rule extraction system," Expert Systems with Applications, vol. 8, 1995.

[13] D. T. Pham and M. S. Aksoy, "An algorithm for automatic rule induction," Artificial Intelligence in Engineering, vol. 8, 1994.